%% file: main.tex
\documentclass[10pt,twocolumn,letterpaper]{article}

\usepackage{cvpr}              % To produce the CAMERA-READY version
\input{math_commands.tex}

\usepackage{url}
\usepackage{booktabs}
\usepackage{utfsym}
\usepackage{amsmath}
\usepackage{bbding}
\usepackage{amsfonts}
\usepackage{amssymb}
\usepackage{graphicx}
\usepackage{textcomp}
\usepackage{xcolor}
\usepackage{wrapfig}
\usepackage{color}
\usepackage{multirow}
\usepackage{pifont}
\usepackage{colortbl} 
\usepackage{tikz}
\usepackage{xcolor}
\usepackage{footnote}
\usepackage{tablefootnote}
\usepackage{graphicx}
\usepackage{wrapfig}
\usepackage{xcolor}
\definecolor{mydarkgreen}{RGB}{0,100,0} % Define dark green color

\usepackage{capt-of,etoolbox}


\definecolor{cvprblue}{rgb}{0.21,0.49,0.74}
\usepackage[pagebackref,breaklinks,colorlinks,allcolors=cvprblue]{hyperref}

\def\paperID{14571} % *** Enter the Paper ID here
\def\confName{CVPR}
\def\confYear{2025}

\title{The Photographer's Eye: 
\\
Teaching Multimodal Large Language Models to See and Critique like Photographers
}
\author{
Daiqing Qi\textsuperscript{1,2}, Handong Zhao\textsuperscript{2}, Jing Shi\textsuperscript{2}, Simon Jenni\textsuperscript{2}, Yifei Fan\textsuperscript{2} \\
Franck Dernoncourt\textsuperscript{2}, Scott Cohen\textsuperscript{2}, Sheng Li\textsuperscript{1} 
\vspace{2mm}
\\
\textsuperscript{1}University of Virginia, \textsuperscript{2}Adobe
\vspace{-6mm}
}

\begin{document}

\makeatletter
\let\@oldmaketitle\@maketitle% Store \@maketitle
\renewcommand{\@maketitle}{\@oldmaketitle% Update \@maketitle to insert...
  \centering
  \includegraphics[width=0.96\textwidth,height=0.4\textheight,keepaspectratio]{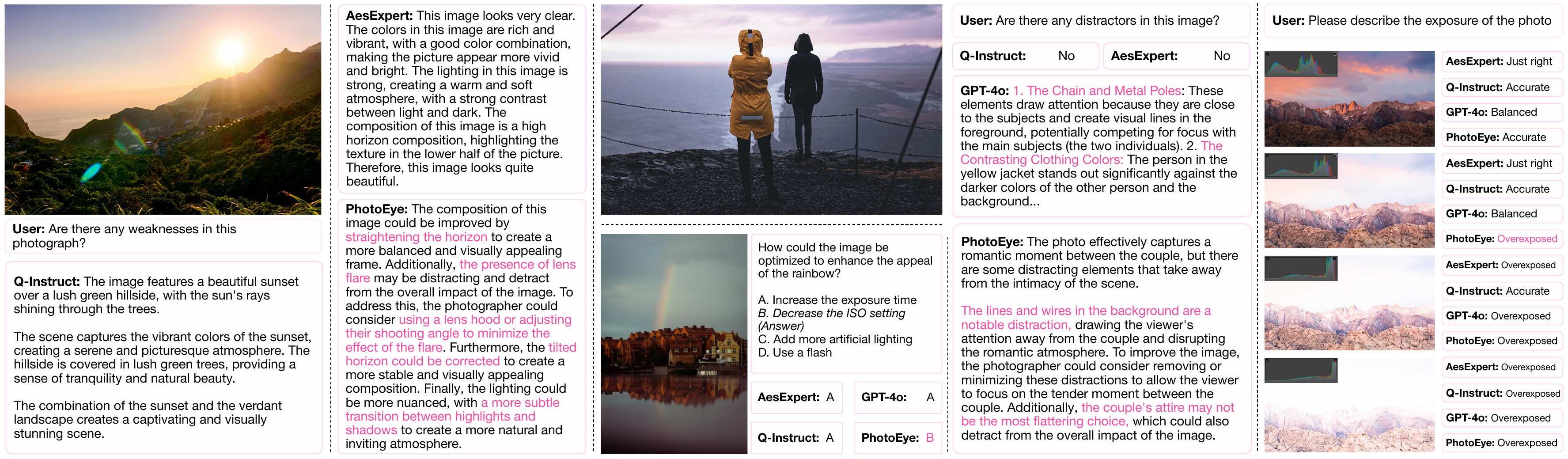}\bigskip
  \vspace{-5mm}
  \captionsetup{width=0.93\textwidth} % 设置caption的宽度
  \captionof{figure}{
  Examples on our model (PhotoEye), existing MLLMs tailored for low-level vision or aesthetics, and GPT-4o \textit{(2024-08-06)}.
The left example and the middle top example highlight a notable limitation of existing open-source MLLMs for low-level vision and aesthetics: \textbf{insufficient coverage of visual aesthetics}. When they fail to identify issues, they either provide positive aspects or claim there are no issues, significantly limiting their usefulness in real-world scenarios. 
The middle bottom example shows existing MLLMs' lack of expertise in photography: lowering exposure properly can instead enhance colors of objects, so B is correct. While other models, including GPT-4o, made mistakes, our model is correct. The right example reveals another clear limitation of existing open-source MLLMs: their vision encoders are \textbf{insensitive to low-level vision and aesthetics}. 
In a series of increasingly overexposed photos (2-nd, 3-rd, and 4-th), PhotoEye’s vision modules, more attuned to low-level and aesthetic features, identified overexposure by the 2-nd photo, while other models recognized it only when the photos were severely overexposed (i.e., 3-rd and 4-th images). 
High-quality aesthetic content is \textcolor[HTML]{f04ea2}{\textbf{highlighted}}.
  }
  \label{fig:head}
  \bigskip
  }% ... an image
\makeatother

\maketitle
\input{sec/0_abstract}
\input{sec/1_intro}
\input{sec/2_related_works}
\input{sec/3_dataset}
\input{sec/4_model}

% \input{sec/5_benchmark}
\input{sec/6_exp.tex}
\input{sec/7_conclusion.tex}
\section*{Acknowledgment}
This work was partially supported by gifts from Adobe Research and the National Science Foundation under Grants IIS-2316306 and CNS-2330215.
{
    \small
    \bibliographystyle{ieeenat_fullname}
    \bibliography{main}
}


\end{document}

%% file: math_commands.tex
\usepackage{amsmath,amsfonts,bm}

\def\eqref#1{equation~\ref{#1}}
\def\1{\bm{1}}

\def\rmF{{\mathbf{F}}}

\def\rmO{{\mathbf{O}}}

\def\rmQ{{\mathbf{Q}}}

\def\rmT{{\mathbf{T}}}

\def\rmX{{\mathbf{X}}}

\def\vw{{\bm{w}}}

\def\mW{{\bm{W}}}

\DeclareMathAlphabet{\mathsfit}{\encodingdefault}{\sfdefault}{m}{sl}
\SetMathAlphabet{\mathsfit}{bold}{\encodingdefault}{\sfdefault}{bx}{n}

%% file: sec/0_abstract.tex
\renewcommand{\thefootnote}{}
\footnotetext{Part of this work was completed during Daiqing's internship at Adobe.}
\renewcommand{\thefootnote}{\arabic{footnote}}

\begin{abstract}

% Photographer and curator 
% , and former director of photography at the Museum of Modern Art (MoMA), 
\vspace{-5mm}
\begin{center}
    \begin{quote}
        \leftskip= -0.5em 
        \rightskip= -0.5em 
        \textit{``While editing directly from life, photographers have found it too difficult to see simultaneously both the blue and the sky.''}
        \vspace{-0.8em} 
        \begin{flushright}
            John Szarkowski, \textit{William Eggleston’s Guide \footnote{William Eggleston is an American photographer, widely recognized as a pioneering figure in color photography.}}
        \end{flushright}
    \end{quote}
\end{center}
% \vspace{-1mm}

% John Szarkowski said in \textit{William Eggleston’s Guide}
% \footnote{William Eggleston is an American photographer, widely recognized as a pioneering figure in color photography.}, \textit{``While editing directly from life, photographers have found it too difficult to see simultaneously both the blue and the sky"}.
%
\noindent Photographer and curator, Szarkowski insightfully revealed one of the notable gaps between general and aesthetic visual understanding: while the former focuses on identifying the factual element in an image (sky), the latter transcends such object identification, viewing it instead as an aesthetic component—a pure color block (blue).
% a pure expanse of blue, appreciated for its contribution to visual aesthetics purely as a color block.
%
Such fundamental distinctions between general (detection, localization, etc.) and aesthetic (color, lighting, composition, etc.) visual understanding present a significant challenge for Multimodal Large Language Models (MLLMs).
% in understanding image aesthetics.
%
% limitations
%
Although some recent works have made initial explorations, they are often limited to general and basic aesthetic commonsense. As a result, they frequently fall short in real-world scenarios (Fig.~\ref{fig:head}), which require extensive expertise—including photographic techniques, photo pre/post-processing knowledge, and more, to provide a detailed analysis and description.
To fundamentally enhance the aesthetics understanding of MLLMs, we first introduce a \textbf{novel dataset}, PhotoCritique, derived from extensive discussions among professional photographers and enthusiasts, and characterized by the large scale, expertise, and diversity.
% characterized by (1) scale: a collection with 20 times the images of existing works, (2) expertise: insights from extensive discussions by photographers and enthusiasts, and (3) diversity: a wide range of photo types and aesthetic perspectives. 
%
Then, to better learn visual aesthetics from PhotoCritique, we furthur propose a \textbf{novel model}, PhotoEye, featuring a language-guided multi-view vision fusion mechanism to understand image aesthetics from multiple perspectives. 
Finally, we present a \textbf{novel benchmark}, PhotoBench, a comprehensive and professional benchmark for aesthetic visual understanding.
On existing benchmarks and PhotoBench, our model demonstrates clear advantages over existing models. Resources available at \href{https://github.com/daiqing98/The-Photographers-Eye}{https://github.com/daiqing98/The-Photographers-Eye}.

\end{abstract}

%% file: sec/1_intro.tex
\vspace{-4mm}
\section{Introduction}
\label{sec:intro}

% general introduction

% \begin{center}
%     \begin{quote}
%         \textit{``To them, composition, light, shade, form and texture are so many catch phrases...''}
%     \end{quote}
% \end{center}

% With increasing attention and significant investments from academia and industry, Multimodal Large Language Models (MLLMs)~\cite{liu2024visual, OpenAI2023GPT4} have achieved great success in general visual tasks. However, one critical capability in human visual perception, which is highly desired in various real-world applications such as image recommendation, editing, generation, etc., remains largely unexplored: visual aesthetics perception. Despite their advancements in general-purpose tasks, aesthetic visual understanding remains particularly challenging.
% The primary limitation of current MLLMs in assessing image aesthetics stems from the fundamental distinction between general and aesthetic visual understanding. While general visual understanding primarily focuses on the factual elements in images—such as object recognition, detection, or segmentation—aesthetic visual understanding requires a more nuanced perception of aesthetic elements, such as chromatic properties (hue, contrast, and saturation), low-level image features, and abstract compositional principles. The disparity between these two modes of visual perception presents a significant challenge for existing MLLMs in accurately evaluating and interpreting image aesthetics.
% We attribute the shortcomings of existing MLLMs in aesthetic visual understanding to two primary factors: dataset and model.

Multimodal Large Language Models (MLLMs)~\cite{liu2024visual, OpenAI2023GPT4} have succeed in general visual tasks but still struggle with a critical ability in human visual perception: aesthetic visual understanding, which is desired in various applications like image recommendation, editing, or generation. The challenge lies in the fundamental distinction between general and aesthetic visual understanding: while the former focuses on factual elements like identification of objects, the latter requires nuanced perception of aesthetic elements such as chromatic properties, or compositional principles. 
We attribute the shortcomings of MLLMs in aesthetic understanding to two primary factors: dataset and model.

% \textbf{Dataset:} Current datasets such as Q-instruct~\cite{wu2024qinstruct} and AesExpert~\cite{huang2024aesexpert}, include content related to low-level vision or visual aesthetics. However, they are limited by (1) data scale, (2) expertise, and (3) diversity. These limitations lead to models showing clear deficiencies in understanding image aesthetics in \textit{real-world scenarios}, where images vary significantly in categories such as shooting conditions and photographic expression.
% %
% Specifically, recent works~\cite{wu2024qinstruct, huang2024aesexpert} typically collect around 20K images from different datasets with around 58K to 88K human annotations. 
% %
% These datasets face two major limitations: first, the data scale is small in terms of both images and annotations. Comprehensive visual aesthetics understanding requires a large volume of training images, as real-world photos encompass diverse subjects, scenes, expression styles, environments, photographic techniques, and so on. A dataset of 20K images is insufficient to cover this range. Secondly, annotations from existing works~\cite{wu2024qinstruct, huang2024aesexpert} for each image come from a fixed, small group of annotators, which notably limits both the diversity and expertise of the annotations. Although the annotators are reportedly trained, their expertise remains limited numerous professional photographers and enthusiasts.
% %

\noindent \textbf{Dataset:}
Current datasets such as Q-Instruct~\cite{wu2024qbench} and AesExpert~\cite{huang2024aesbench}, made initial explorations in low-level vision or visual aesthetics with MLLM but suffer from limitations in scale, expertise, and diversity. 
With around 20K images and 58K–88K image annotations, these datasets are insufficient for comprehensive aesthetic understanding, showing clear deficiencies in real-world scenarios, where images vary widely in subjects, scenes, styles, environments, and techniques, requiring larger datasets to ensure coverage. Additionally, their annotations come from small (typically tens of annotators), fixed groups of annotators, limiting diversity and expertise. Even if the annotators are reportedly trained, their expertise remains limited compared to professional photographers and enthusiasts.

\noindent \textbf{Model:} When adapting to aesthetic visual understanding, existing approaches~\cite{wu2024qinstruct, huang2024aesexpert, zhou2024uniaa, huang2024visualcritic} typically finetune existing MLLMs like LLaVA~\cite{liu2024visual} on customized datasets, ignoring the unique requirements of visual features for aesthetic visual understanding. Comprehensive understanding of image aesthetics requires multi-view information: for example, high-level features are essential for interpreting narrative and storytelling, while detailed spatial information is critical for understanding composition. Similarly, low-level details are required for  perception of color and lighting, etc.

To address the challenges, we first introduce the \textbf{PhotoCritique} dataset, featuring (1) large scale: containing over \textbf{450K} images with detailed aesthetic descriptions, and around \textbf{20} times the size of existing datasets~\cite{wu2024qinstruct, huang2024aesexpert}; and (2) expertise and diversity: the aesthetic descriptions are derived from extensive discussions among large groups of professional photographers and enthusiasts online, rather than a fixed small number of annotators as in existing works~\cite{wu2024qinstruct, huang2024aesexpert}. 
Subsequently, given the notable different focuses in general and aesthetic visual understanding, we further introduce \textbf{PhotoEye}, an MLLM tailored for aesthetic visual understanding. PhotoEye features a language-guided multi-view vision fusion mechanism to effectively learn aesthetic knowledge from PhotoCritique, enabling it to understand image aesthetics from multiple perspectives for various aesthetics-related questions.

Finally, to provide a comprehensive evaluation of our model and existing MLLMs on aesthetic visual understanding, we introduce our new benchmark \textbf{PhotoBench}. PhotoBench draws its questions from in-depth discussions among professional photographers and enthusiasts, covering 284 sub-topics in photography, making it notably more professional and diverse than existing works. 
We summarize our contributions as follows:

\begin{itemize}
    \item We introduce PhotoCritique, the first large-scale, professional, and diverse dataset for aesthetic visual understanding, consisting of 450K images with 2.63M instruction-tuning pairs, sourced from extensive discussions among professional photographers and enthusiasts.
    
    \item We propose PhotoEye, an MLLM specifically designed for aesthetic understanding, featuring a language-guided multi-view vision fusion mechanism.
    
    \item We develop PhotoBench, a professional  benchmark for aesthetic visual understanding, offering 284 photography sub-topics and insights from photography professionals.

    \item We conduct a comprehensive study on aesthetic visual understanding with our model and existing MLLMs, revealing the limitations of existing MLLMs and demonstrating how we overcome them through extensive experiments.
    
\end{itemize}

%% file: sec/2_related_works.tex
\section{Related Works}
\label{sec:formatting}
\noindent \textbf{Multimodal Large Language Models.} 
MLLMs~\cite{chen2023minigpt, liu2024visual} are evolving rapidly recently, demonstrating advanced capabilities in image-text dialogue through enhanced alignment and fine-tuning. Further studies \cite{chen2023internvl, bai2023qwen, dai2023instructblip, ye2023mplug, tong2024cambrian1fullyopenvisioncentric} have strengthened MLLMs by increasing the data scale or improving the backbone model. 
Furthermore, with customized data, a branch of works \cite{ye2023mplug, you2023ferret, chen2023shikra, peng2023kosmos} extend MLLMs' capabilities to referring and grounding. However, despite these advancements in general visual tasks, the aesthetic visual understanding of MLLMs remains largely underexplored. 

\noindent \textbf{Understanding Image Aesthetics with MLLM.}
Traditional Image Aesthetics Assessment (IAA)~\cite{murray2012ava, yang2022personalized, ren2017personalized} is often formalized as a regression task, where the model learns to predict the score of the aesthetic quality of images. One notable limitation is the lack of interpretability, which is critical in practice because we also want to understand why the image is good or bad. Recent advances in MLLM make it possible: a few recent works~\cite{zhou2024uniaa, wu2024qbench, wu2024qinstruct, huang2024aesbench, huang2024aesexpert, huang2024visualcritic, wu2024openended} explore \textit{interpretable} aesthetic understanding with MLLMs. However, they often struggle with the limited data scale or lack diverse, professionally curated aesthetic descriptions.

\noindent \textbf{Improving Visual Perception in MLLM.}
Recently, several studies~\cite{Zong2024, Lin2023, tong2024cambrian1fullyopenvisioncentric, qi2024tag, qi2024easy, qi2024generalizing} have investigated backbone vision modules for MLLM. 
% Specifically, they extend the single CLIP vision encoder in MLLMs to multiple encoders, enabling the extraction of richer visual features. 
While SPHINX~\cite{lin2024sphinx} simply concatenates features from different encoders, Cambrian~\cite{tong2024cambrian1fullyopenvisioncentric} and MoVA~\cite{Zong2024} merge different visual features into a single one. However, the fusion process in MoVA is based on the CLIP feature, which may introduce bias, and the feature extraction (prior to fusion) lacks language-based control.
Cambrian~\cite{tong2024cambrian1fullyopenvisioncentric} employs query-based attention, 
however, visual features from different encoders are simply averaged during fusion, and the queries are independent of language instructions.
To address these limitations, we propose a new MLLM, PhotoEye, featuring a language-guided multi-view vision fusion mechanism to effectively learn aesthetic knowledge.

%% file: sec/3_dataset.tex
\section{Learning with Photography Enthusiasts and Professionals}

One key reason existing MLLMs struggle with aesthetic visual understanding in real-world scenarios is the training data limitations: (1) Data Scale: Current datasets~\cite{wu2024qinstruct, huang2024aesexpert} ($\sim$20K images) lack sufficient coverage of photography genres, styles, themes, etc. High-quality aesthetic descriptions are also limited, with typically 57K-88K detailed descriptions in existing works~\cite{wu2024qinstruct, huang2024aesexpert, zhou2024uniaa}, while our dataset provides over \textbf{450K} images and detailed descriptions. (2) Description Quality: Most datasets~\cite{wu2024qinstruct, huang2024aesexpert} rely on a small, fixed group of annotators, limiting expertise and diversity. In contrast, our PhotoCritique dataset draws insights from hundreds of thousands of photographers and enthusiasts, yielding more professional, nuanced feedback.

\begin{figure}[t]
\vspace{-1mm}
\centerline{\includegraphics[scale=0.54]{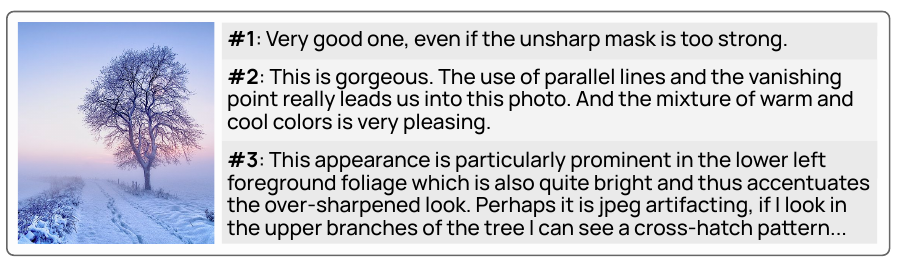}}
\caption{An example from the dpchallenge platform. 3 of the total 64 comments for the photo are presented for illustration.}
\label{fig:dpc}
\vspace{-4mm}
\end{figure}

\subsection{Mining from Digital Photography Platforms}
% \vspace{1.5mm}
% \noindent \textbf{Mining from Digital Photography Platforms.}
 \noindent To address the limitations, we avoided using annotators and instead drew insights from a broad range of online communities.
The Digital Photography Challenge (DPC) is an online platform, where users submit their works across various themes and genres and receive constructive feedback from other photographers or enthusiasts (Fig.~\ref{fig:dpc}).
The wealth of high-quality discussions paired with images on DPC provides valuable resources for MLLMs to learn visual aesthetics. However, directly using this data for supervised fine-tuning is impractical due to the significant noise present in the raw comments.
To address this issue, we developed strategies to analyze, summarize, and refine raw comments into a single, well-crafted critique for each image.
\begin{figure}[t]
% \vspace{-1mm}
\centerline{\includegraphics[scale=0.53]{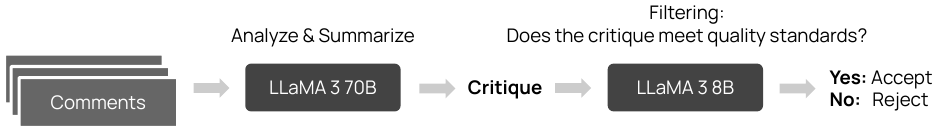}}
\caption{Data Generation Pipeline. The generation of one critique from a group of comments of one image is shown as an example.}
\label{fig:data_pipe}
\vspace{-2mm}
\end{figure}
First, we need to collect raw data from \textit{dpchallenge.com} online. Since existing works~\cite{zhong2023aesthetically} already collected raw data prior to 2022 from the website, we can skip the data scraping process and focus on data processing. 
Given a post from the DPC platform (comprising one uploaded image and a series of user comments, shown in Fig.~\ref{fig:dpc}), we design a multi-step pipeline to generate a single unified photo critique using prompt-based methods from \textbf{comments only} without the image. 
Fig.~\ref{fig:data_pipe} briefly illustrates the pipeline with the example of generating a single critique from a set of comments of one image. 
we first instruct the LLM to analyze and summarize the raw comments, Then the LLM is asked to integrate the comments into a cohesive and unified photo critique. While LLMs possess strong reasoning and comprehension abilities, the immediate output critique is sometimes sparse in information or misleading. 
Therefore, we designed a filtering strategy to eliminate low-quality data with a smaller LLM for improved efficiency. During the filtering stage, generated photo critiques are input into the smaller LLM, and prompt-based strategies are used to detect whether the generated photo critique conveys enough aesthetics-related information. Only critiques that pass this filtering stage are accepted. We term this portion of data as\textit{ aesthetic description}.
Details available in Appendix A.

\begin{figure}[t]
\vspace{-2mm}
\centerline{\includegraphics[scale=0.25]{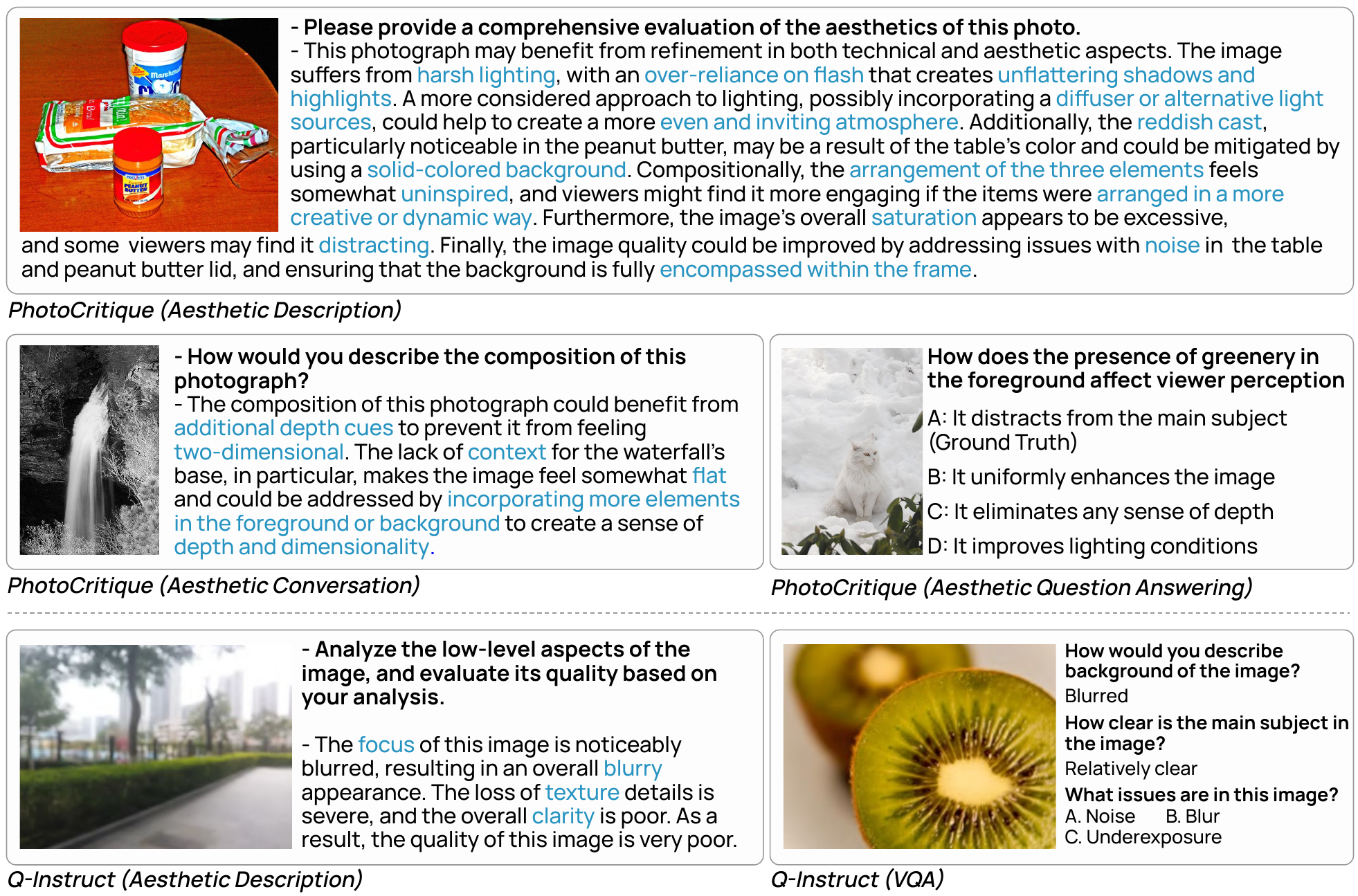}}
\caption{\textbf{Top:} Examples of PhotoCritique. \textbf{Bottom:} Comparison with examples in Q-Instruct~\cite{wu2024qinstruct}.}
\label{fig:pc}
\vspace{-6mm}
\end{figure}

\begin{figure}[t]
\vspace{-2mm}
\centerline{\includegraphics[scale=0.35]{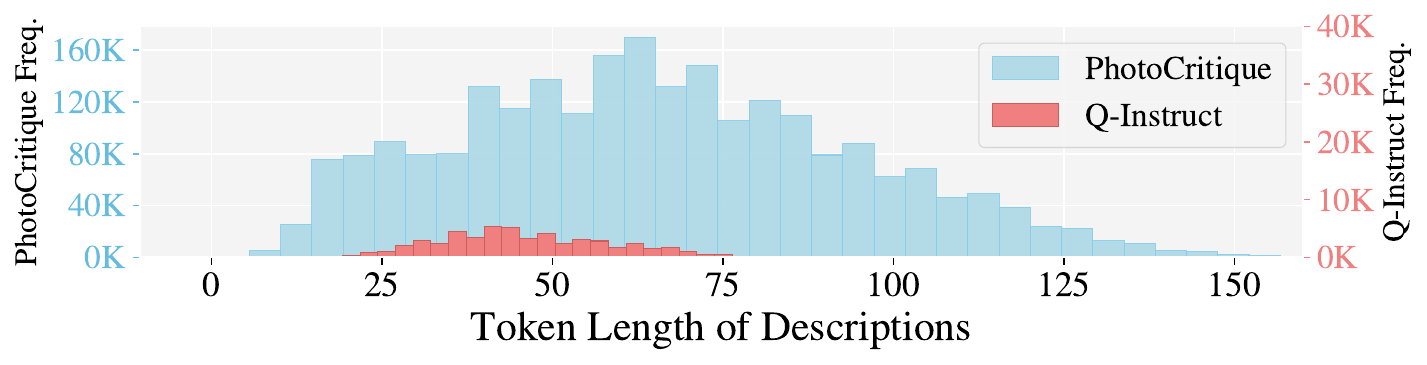}}
\vspace{-2mm}
\caption{Comparison of description length Distributions of aesthetic comments in PhotoCritique and Q-Instruct. Scale of PhotoCritique is \textcolor[HTML]{87CEEB}{Blue} and that of Q-Instruct is \textcolor[HTML]{F08080}{Red}.}
\label{fig:len_dis}
\vspace{-2mm}
\end{figure}

% \subsection{From Photo Critiques to Diverse Instruction-tuning Pairs}
\vspace{1.5mm}
\noindent \textbf{From General Photo Critiques to Diverse Instruction-tuning Pairs.}
\noindent In the previous step, we generated photo critiques from DPC’s raw comments, which provide overall aesthetic reviews of the given images. 
In this step, we further utilize LLMs to generate multiple high-quality, diverse conversations addressing various aspects of image aesthetics, such as lighting, composition, color, emotion, narrative, and so on, as well as extensive photographic techniques and pre/post-processing skills. 
Another critical reason for this step is that raw comments are sometimes too lengthy, which can lead to certain aesthetic aspects being overlooked in the LLM-generated critique. By specifically prompting about these aesthetic aspects, the LLM can effectively summarize essential information related to each aspect from the lengthy raw comments.
Similarly, after generating all the question-answer pairs, we apply prompt-based filtering strategies to reject less informative pairs, accepting only the high-quality samples. We term this portion of data as \textit{aesthetic conversation}.
Additionally, we also generate MCQs from photo critiques with LLMs. Specifically, each critique is input into the LLM with prompts to generate five MCQs based on the ground truth provided in it. We term this portion of data as \textit{aesthetic VQA}.
\textcolor{black}{Details available in Appendix A}.

\subsection{PhotoCritique}

\begin{figure}[t]
% \vspace{-2mm}
\begin{subfigure}{\linewidth}
   \centerline{\includegraphics[scale=0.20]{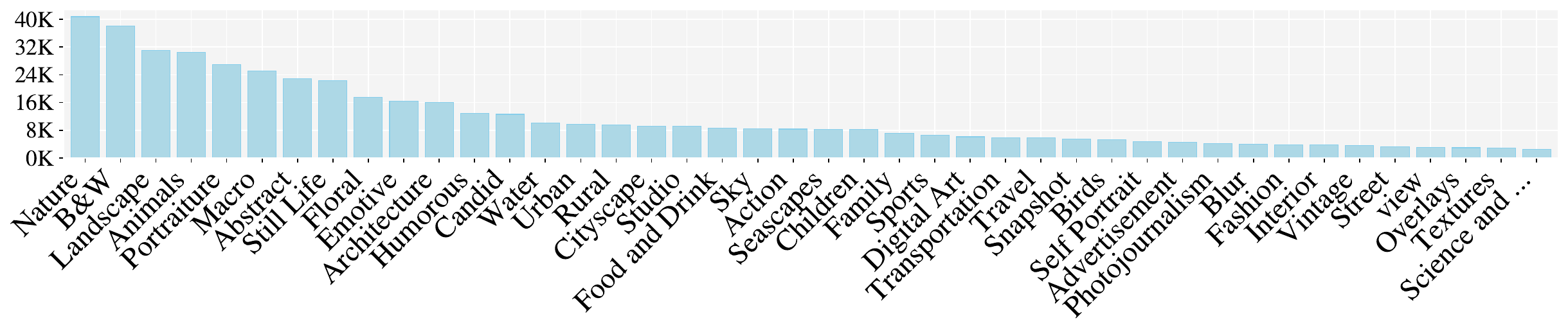}}
\end{subfigure}

\begin{subfigure}{\linewidth}
   \centerline{\includegraphics[scale=0.26]{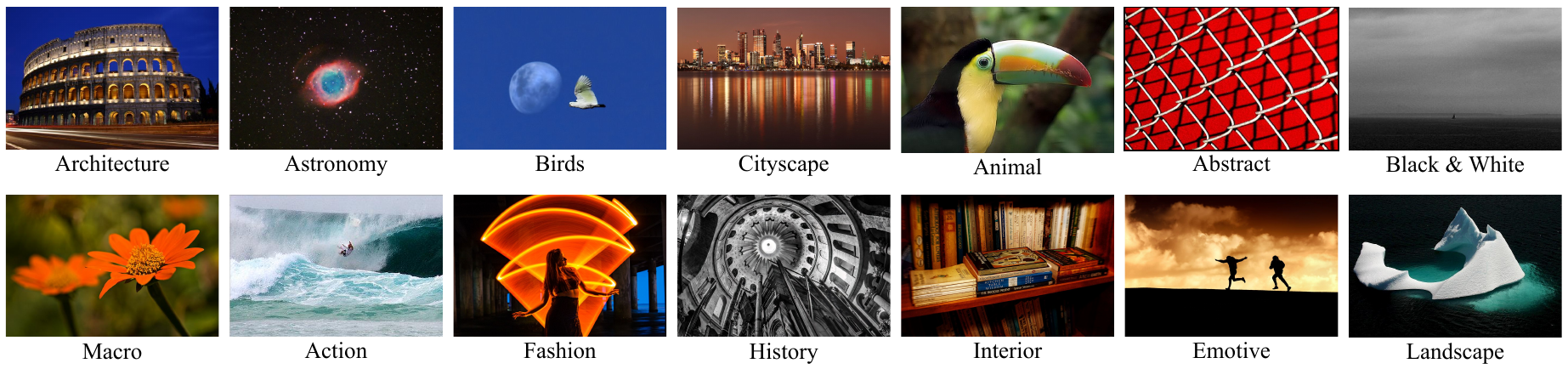}}
\end{subfigure}

\caption{\textbf{Top:} Distribution of Top 40 categories. \textbf{Bottom:} Examples of photos from a subset of categories.}
\label{fig:key_dis}
\vspace{-7mm}
\end{figure}

\noindent \textbf{Overview.} Combining all above, we build PhotoCritique, the first large-scale dataset for expert-level image aesthetics understanding.
PhotoCritique consists of three parts: aesthetic description, aesthetic conversation, and aesthetic VQA. Examples of each are shown in Fig.~\ref{fig:pc} (Top). The three parts contain 450K, 1.9M, and 250K samples, respectively.
% as illustrated in Fig.~\ref{fig:pc_pie}.
% Finally, we combined all the data above to build the PhotoCritique, the first large-scale dataset for expert-level image aesthetic understanding.
Born from a large volume of shared images and discussions by photography enthusiasts and professionals on online photography forums, PhotoCritique covers a wide range of over 70 photographic categories, encompassing diverse scenes and subjects. Fig.~\ref{fig:key_dis} shows the distribution of the top 40 categories along with several image examples. The extensive scale and diversity enhance the model's effectiveness in real-world scenarios.

% -------------------------------------------------
\begin{table}[t]
\vspace{-0mm}
\centering
\scalebox{0.7}{ 
\small
\setlength{\tabcolsep}{6pt}  % col
\renewcommand{\arraystretch}{0.8}  % row
\begin{tabular}{lcccc}
\toprule
\textbf{Dataset} & \textbf{\#Image} &\textbf{ \#Annotations } & \textbf{\#Training Samples} & \textbf{\#Avg. Length} \\
\midrule
UNIAA & 57K & 57K & 57K & 32.8 \\
AesExpert* & 21K & 88K & - & - \\
Q-Instruct & 18K & 58K & 66K & 46.4 \\
\midrule
PhotoCritique& \textbf{450K} & \textbf{450K} &\textbf{2.4M} & \textbf{65.2}  \\
\bottomrule
\end{tabular}
}
\caption{Statistics of samples attributed to \textit{detailed aesthetic descriptions and conversations} across different datasets. Datasets with * are currently not available.}
\label{tab:datasets}
\vspace{-6mm}
\end{table}
% -------------------------------------------------

\begin{figure*}[t]
\vspace{-2mm}
\centerline{\includegraphics[scale=0.23]{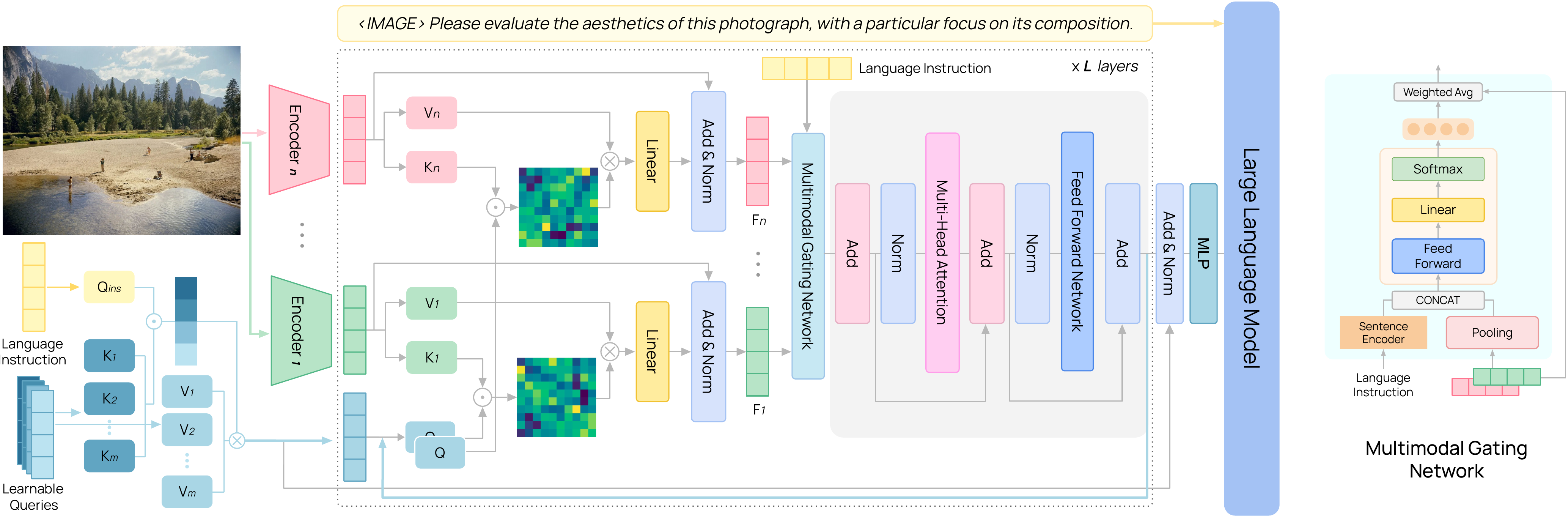}}
\caption{Framework of PhotoEye. Given an input photo and a language instruction, visual features from different visual encoders are fused in our vision fusor with language guidance, then fed into the LLM with language embeddings to generate LLM response.}
\label{fig:fw}
\vspace{-2mm}
\end{figure*}

\vspace{1.5mm}
\noindent \textbf{Comparison with Existing Datasets.} 
PhotoCritique represents a fundamental improvement over existing works in both quantity and quality.
\textbf{Data Scale:} 
we compares PhotoCritique with existing datasets by showing statistics for data related to detailed aesthetic descriptions and conversations in Tab.~\ref{tab:datasets}. This includes the number of source images, the annotations for these images, and the final count of instruction-tuning pairs generated from these annotations using LLMs. Additionally, we calculate the average length of aesthetic comments (including both descriptions and conversations) in the training data. Tab.~\ref{tab:datasets} focuses on training data statistics for aesthetic comments, as these are essential in practical applications where users need detailed feedback on why their photos are strong or weak and how to improve them. In contrast, Multiple Choice Questions (MCQ) are relatively limited in this regard. Besides, aesthetic comments better reflect the information density of the dataset than MCQs, as generating MCQs from a detailed comment is easy and can result in numerous MCQs of varying quality depending on the input prompt. Therefore, the number of MCQs serves only as a relatively rough indicator of the information density, although the number of MCQs in PhotoCritique is also the largest.
As shown in Tab.~\ref{tab:datasets}, PhotoCritique demonstrates a notable improvement across all metrics compared to existing datasets.
\noindent \textbf{Data Quality:} 
PhotoCritique is more professional, and diverse. Unlike most existing works~\cite{wu2024qinstruct, huang2024aesexpert}, which rely on a small group (usually a few dozen) of annotators to label data, our dataset is sourced from images and discussion from over 107,000 photography enthusiasts and professionals on online photography forums. 
Fig.~\ref{fig:dpc} provides an intuitive comparison between our dataset and Q-Instruct, with examples of Q-Instruct in its original paper. While Q-Instruct primarily focuses on basic low-level visual facts, PhotoCritique not only addresses low-level details but also considers how these elements impact image aesthetics and ways to improve them. Additionally, we include discussions on photography techniques (e.g., the first example in the aesthetic description covers flash usage and lighting) and the overall impression conveyed by the photo. 
The comprehensiveness of aesthetic comments in PhotoCritique is also reflected in the average length: Fig.~\ref{fig:len_dis} shows the distribution of aesthetic comment lengths in PhotoCritique and Q-Instruct, with PhotoCritique containing a large number of detailed descriptions. Tab.~\ref{tab:datasets} also shows that the average length of aesthetic comments in PhotoCritique is notably greater than that of existing datasets.
\textcolor{black}{More examples in Appendix A.}

%% file: sec/4_model.tex
\section{Improving Visual Aesthetics Perception}

\subsection{See like Photographers}
\begin{figure}[t]
\vspace{-0mm}
\centerline{\includegraphics[scale=0.23]{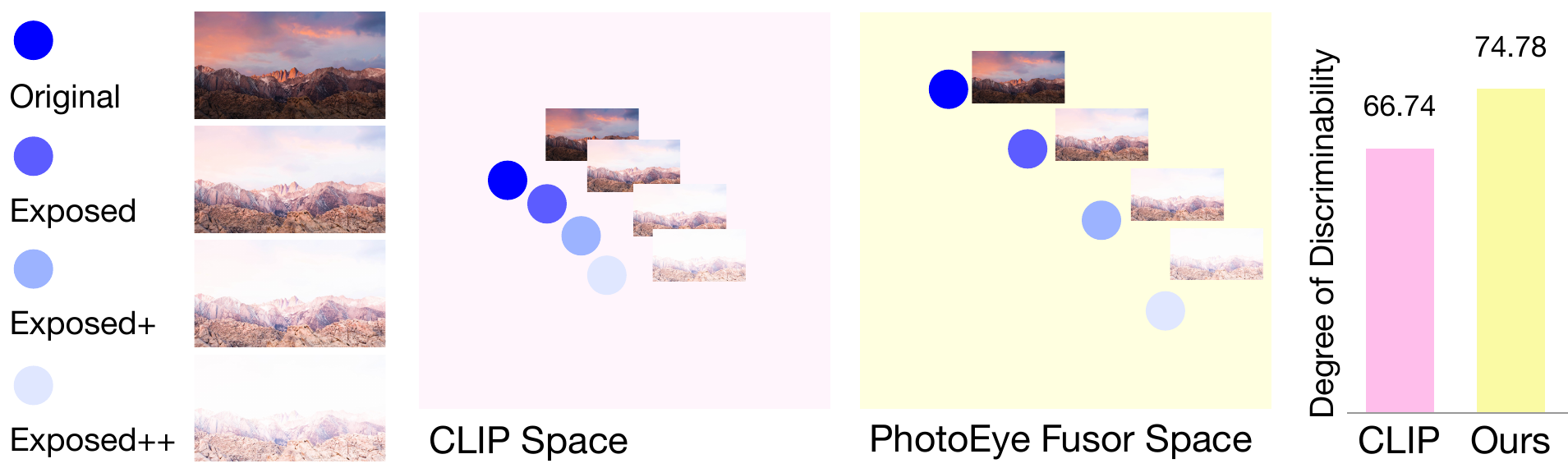}}
\caption{Comparison of aesthetics-related visual feature discriminability of ours and CLIP~\cite{liu2023improved} in existing works~\cite{huang2024aesexpert, wu2024qinstruct, zhou2024uniaa}.}
\label{fig:eye_intro}
\vspace{-6mm}
\end{figure}

In addition to limitations in existing datasets, Open-source MLLMs face another notable challenge: the limited sensitivity of the vision encoder to aesthetic elements.
Most existing works~\cite{huang2024aesexpert, wu2024qinstruct} use CLIP as the vision encoder, which, however, is pre-trained with high-level image-text alignment in general domains and is relatively less effective at capturing aesthetic elements. Fig.~\ref{fig:head} (right) illustrates this issue, where Q-Instruct and AesExpert only report `overexposure' when the image is severely overexposed. In Fig.~\ref{fig:eye_intro}, we quantitatively validate it by calculating the discriminability (defined as the average distance between these embeddings) of aesthetic features from different vision encoders. \textcolor{black}{Details are available in Appendix B.} It shows that CLIP is less effective in discriminating towards aesthetics-related features as they are more densely clustered.
% Specifically, we pick up a set of common factors in photography such as exposure, then sample 100 images from RPCD~\cite{zhong2023aesthetically} and for each image, systematically shift each factor to different degrees. Then we compare the discriminability of them in embedding space of different encoders.
To handle the diversity of aesthetics understanding, from high-level elements like emotion and storytelling to low-level features such as lighting, color, and composition, we propose PhotoEye, a MLLM with a language-guided multi-view vision fusor, which enhances visual perception by fusing multiple vision encoders pre-trained on complementary tasks.

\subsection{PhotoEye} 
\noindent \textbf{Architecture.}
PhotoEye is shown in Fig.~\ref{fig:fw}. It consists of a multi-view vision fusor and a backbone LLM.
Given an image and an aesthetics-related language instruction, visual features of the image from different vision encoders are fused into a single visual feature by the multi-view vision fusor, conditioned on both the image and the \textit{language instruction}.
Specifically, the multi-view vision fusor consists of $L$ fusion blocks. In each $l$-th block (or layer), a group of visual features $\{\rmX^{i}  \in \mathbb{R}^{C \times H \times W} \}_{i=1}^{N}$ from $N$ different vision encoders are fused with a learnable query $\rmQ^{l} \in \mathbb{R}^{C \times H \times W}$. Denote the output of the $l$-th block as $\rmO^{l}$, $\rmQ^{l+1}=\rmO^{l}, \text{ if } l \neq 0$. The first learnable query is generated by the language-guided query generator.

\vspace{1.5mm}
\noindent \textbf{Language-guided Query Generation.}
At the first layer, $\rmQ^{1}$ is generated by the language-guided query generator (bottom left in Fig.~\ref{fig:fw}). 
Specifically, we maintain a group of $M$ learnable queries  $\{\rmQ^{i}  \in \mathbb{R}^{C \times H \times W} \}_{i=1}^{M}$. Given a specific language instruction, it is encoded into a text embedding $\rmT$, which serves as the query to dynamically generate a single query from all $M$ learnable queries with attention mechanism: $\rmQ^{1} = \text{ATTN}(\mW_{q}^{\text{T}} \rmT, \mW_{k}^{\text{T}} \rmQ^{}_{\text{M}}, \mW_{v}^{\text{T}} \rmQ^{}_{\text{M}})$, where $\rmQ^{}_{\text{M}}$ is the concatenation of all $M$ queries, and $\mW_{q}^{\text{T}}$, $\mW_{k}^{\text{T}}$, $\mW_{v}^{\text{T}}$ are the projection matrixes of the query, key and value, respectively.
Note that a pre-trained BERT is introduced as the text encoder, where the [CLS] token from the output is used as the text token $\rmT$. We avoid the CLIP text encoder to prevent bias towards the CLIP visual features in the later multimodal gating process.

In this way, for each language instruction that may focus on a specific aesthetic concept, a tailored query is generated from all $M$ learnable queries, which can best extract visual features related to the aesthetic concept from each vision encoder in the fusion block during inference.

%%%%

\vspace{1.5mm}
\noindent \textbf{Feature Extraction with Learnable Query.}
At $l$-th layer, given a query $\rmQ^{l}$ and a group of visual features $\{\rmX_{i}\}_{i=1}^{N}$ from $N$ different vision encoders, each visual feature is first interpolated to the same shape:
$\rmX_{n}^{*}$ = Interpolate($\rmX_{n}$), where $\rmX_{n}^{*}  \in \mathbb{R}^{C \times H \times W}$.
Then, each of the extracted visual features from the corresponding vision encoder is given by 
$\rmF_{n}^{l} = \text{ATTN}( \mW_{q}^{n} \rmQ^{l}, \mW_{k}^{n} \rmX_{n}^{*}, \mW_{v}^{n} \rmX_{n}^{*} ) $, where $\mW_{q}^{n}$, $\mW_{k}^{n}$, $\mW_{v}^{n}$ are the projection matrixes of the query, key and value.

\vspace{1.5mm}
\noindent \textbf{Multimodal Gating Network.} 
The extracted visual features $\{\rmF^{l}_{i} \}_{i=1}^{N}$ are then fused into a single visual feature in the multimodal gating network. 
The multimodal gating network assigns different weights to each extracted visual feature conditioned on both image and language and fuses them into one single feature based on their importance. 
The weights of visual features are directly predicted by an MLP. Specifically, given the language embedding  $\rmT$ and a set of visual features $\{\rmF^{l}_{i} \}_{i=1}^{N}$, weights are given by:
$ \vw^{l} = \text{MLP}(\rmT ; \text{Pooling}(\{\rmF^{l}_{i} \}_{i=1}^{N})) , \text{where } \vw \in \mathbb{R}^{ N \times 1} $, and ; means concatenation.
\textcolor{black}{The fused single feature is obtained by $ \hat{\mathrm{F}}^l = \sum_{i=1}^{N} \vw_i^{l} \cdot \mathrm{F}^l_i  $.}
Then, $ \hat{\mathrm{F}}^l $ is further fed forwarded to a transformer block with residual links. The output of the vision fusor is input to the $(l+1)$-th layer if $l < L$; otherwise, it is fed to an MLP and input to the LLM.

\noindent \textbf{Remark.} Compared to existing works, PhotoEye differs from MoVA~\cite{Zong2024} in removing its potential bias towards CLIP features by fusing visual features with a group of learnable queries. Moreover, a language-guided query generation mechanism is introduced to extract different features towards different language instructions. When compared to Cambrian-1~\cite{tong2024cambrian1fullyopenvisioncentric}, PhotoEye conditions both feature extraction and fusion on language, allowing it to capture more refined, instruction-aware aesthetic visual features. In contrast, Cambrian-1 is independent of instructions, potentially resulting in relatively coarser visual features.

\begin{figure*}[t]
\vspace{-1mm}
\centerline{\includegraphics[scale=0.24]{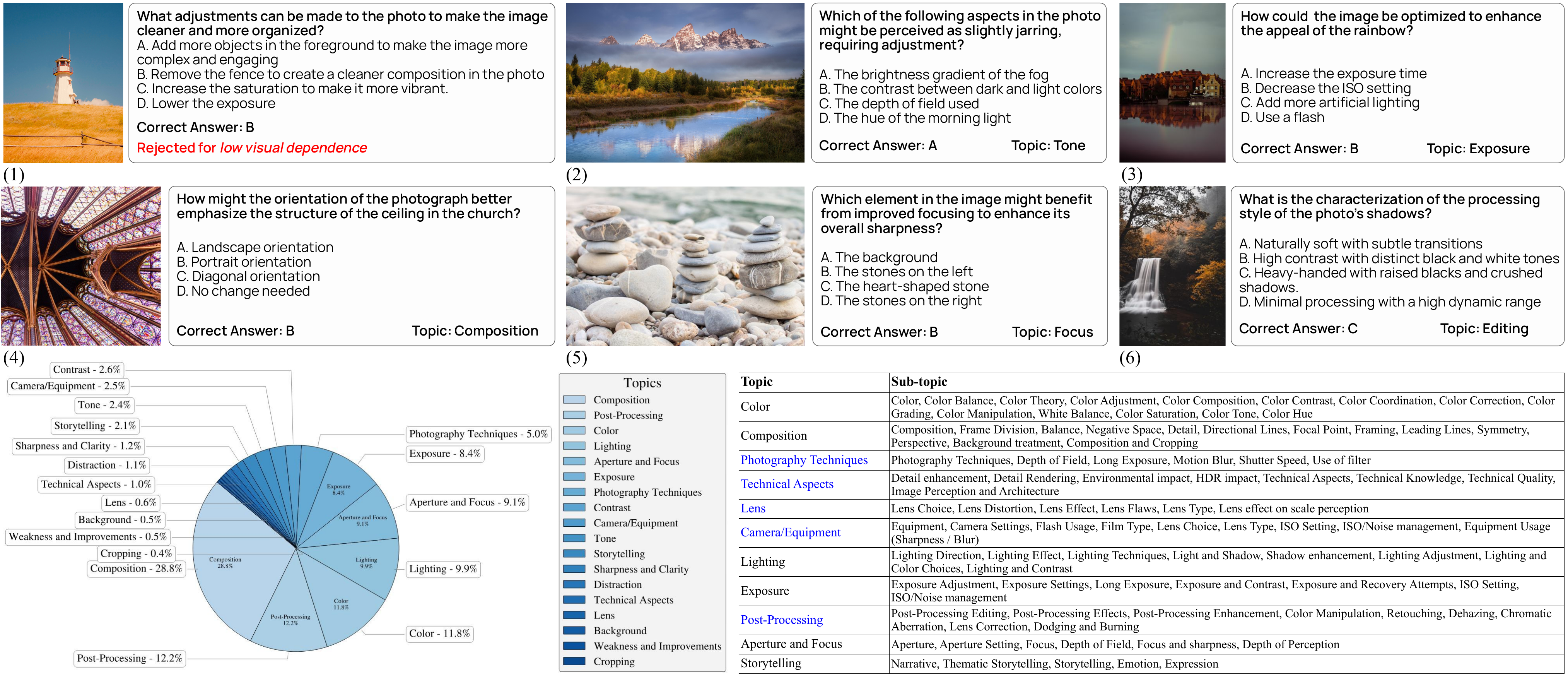}}
\caption{We show one rejected example, several accepted examples, topic distribution, and some sub-topics of a subset of topics of questions in PhotoBench.}
\label{fig:bench}
\vspace{-6mm}
\end{figure*}

% -------------------- -------------------- -------------------- -------------------- --

\section{Benchmarking Aesthetic Visual Understanding with Professionals}

In Section 3, we highlighted a major limitation of existing datasets: annotations are typically provided by a few dozen annotators, which limits the diversity and expertise of the annotations. Even with reported training, these annotators lack the expertise of the many experienced photographers and enthusiasts. Existing benchmarks, such as Q-Bench~\cite{wu2024qbench} and AesBench~\cite{huang2024aesbench}, also suffer from this issue, relying on similar methods to obtain annotations. 

To address the limitation and cover a wider range of scenarios, we directly draw insights from extensive photos and discussions on Reddit’s PhotoCritique, a vibrant photography community.
This approach allows us to generate comprehensive, diverse, and expert-level multiple-choice questions to evaluate the aesthetic visual understanding of MLLMs. Specifically, we do not generate MCQs from annotator-provided annotations; instead, we derive MCQs from photo reviews in Reddit’s PhotoCritique. We refer to our new benchmark as \textbf{PhotoBench}. 
Since RPCD~\cite{nieto2022understanding} has already collected raw data from PhotoCritique posts between 2009 and 2022, we directly design pipelines that use an LLM (GPT-4-turbo) to generate MCQs.

% \vspace{1.5mm}
% \noindent \textbf{From Insights to Questions}

\subsection{From Insights to Questions}
The entire question-generation process consists of three steps: (1) summarization, (2) question generation, and (3) multi-stage filtering. 
\textbf{(1) Summarization.} 
Since the raw comments from the forum are quite noisy, empirically we find directly inputting the raw comments corresponding to an image to the LLM and prompting it to generate multiple-choice questions based on these comments does not yield ideal results. To address this issue, we first use the LLM to summarize the noisy raw comments for each image, producing a single, clean, comprehensive, and detailed photo critique. This process aligns with the pipeline shown in Fig.~\ref{fig:data_pipe}.
\textbf{(2) Question Generation.} For each image, we input its corresponding photo critique into the LLM and instruct it to generate five multiple-choice questions grounded on the provided critique.
\textbf{(3) Multi-stage Filtering.}
This step directly impacts the quality of PhotoBench. We first selected the top 5,000 most detailed photo critiques from the initial generation step and instructed the LLM (using GPT-4 Turbo, as we empirically found it slightly outperforms GPT-4o) to generate 25,000 questions. 
The first filtering stage is LLM-based, primarily aimed at removing MCQs with low visual dependency—that is, questions that could likely be answered correctly by reading only the question and options without viewing the image. We show an example in Fig.~\ref{fig:bench} (1). Our method is intuitive yet effective: we provide only the question and options to the LLM (GPT-4o) without the image and filter out questions it answers correctly. This step eliminated about 70\% of the items. We then instructed GPT-4o to score the remaining questions based on aesthetics relevance, visual dependency, and expertise. 
% We selected the top 3,000 questions by average score and submitted them for review by photographers, who further evaluated them against these standards. After this final review, we obtained 1,500 high-quality MCQs.
\textcolor{black}{We selected the top 1,500 questions based on their average score.}
\textcolor{black}{Details available in Appendix C.}

\subsection{PhotoBench} We show detailed information of PhotoBench in Fig.~\ref{fig:bench}. 
Compared to existing works~\cite{wu2024qbench, huang2024aesbench, zhou2024uniaa}, Photobench gains more advantages in diversity and expertise. 
Due to potentially limited expertise and diversity among annotators~\cite{wu2024qbench, huang2024aesbench}, MCQs from existing works are often too simple. We can have a straightforward comparison in Fig.~\ref{fig:bench_base}.

While questions from existing works are relatively straightforward and focus on factual elements, our questions are more challenging and practical, requiring \textbf{deeper aesthetics understanding} and covering topics \textbf{directly related to photography in practice}, such as camera settings, post-processing, and photographic techniques (topics marked in blue in Fig~\ref{fig:bench}.
% example
Examples are shown in Fig~\ref{fig:bench} (3,6). In Fig.~\ref{fig:bench} (3), to make the rainbow stand out more, the exposure should be reduced (slight underexposure can instead enhance color saturation). In Fig.~\ref{fig:bench} (6), we can observe that the shadows are not purely black but rather "grayish" (zoom in for a better view)—a common post-processing technique achieved by raising the shadow curve to enhance the photo’s texture or create a film-like aesthetic. They suggest that PhotoBench includes more professional and detailed questions, closely aligned with practical applications such as photo post-processing and editing.
%

%% file: sec/6_exp.tex
% \vspace{-2mm}
\section{Experiments}
% \vspace{1mm}
\noindent \textbf{Implementation Details.}
% PhotoEye is finetuned on instruction data for one epoch, following existing works~\cite{liu2023improved, wu2024qinstruct, huang2024aesexpert}. 
% 
We build PhotoEye with Vicuna-v1.5-7B as the LLM backbone, following existing works~\cite{wu2024qinstruct, huang2024aesexpert} for fair comparison.
We also use minimal data from the general domain, including LLaVA-Pretrain~\cite{liu2023improved} for pre-training and LLaVA-665K~\cite{liu2023improved} for instruction finetuning. We mix our data with LLaVA-665K during the finetuning stage.
We use a batch size of 128 and a learning rate of 1$e$-3 for pre-training, and 2$e$-5 for finetuning. The training takes around 96 hours with 8 A100 GPUs with ZeRO2. For vision encoders, we use: CLIP-ViT-L/14~\cite{radford2021learning}, DINOv2-giant~\cite{oquab2023dinov2} CoDETR-ViT-L~\cite{zong2023detrs} and SAM-ViT-H~\cite{kirillov2023segment}.
Details are available in the Appendix D.

\subsection{Results}
We report results of PhotoEye on existing benchmarks. Although PhotoCritique is not specifically designed for low-level vision, but PhotoEye still achieves competing results on Q-Bench when finetuned with it~\cite{wu2024qbench}.
\begin{table*}[t]
    \centering
    \footnotesize
    \renewcommand{\arraystretch}{0.9}    % Row spacing adjustment
    \setlength{\tabcolsep}{12pt}          % Column spacing adjustment
    \scalebox{0.75}{                     % Scale table to 85% of text width
        \begin{tabular}{llcccccccccc}    % Removed @{ } to avoid extra spacing
            \toprule
            \multirow{3}{*}{\textbf{Model}} & \multirow{3}{*}{\textbf{LLM Backbone}} & \multicolumn{3}{c}{\textbf{Question Types}} & \multicolumn{4}{c}{\textbf{Quadrants of Low-level Concerns}} & \multirow{3}{*}{\textbf{Overall↑}} \\
            \cmidrule(lr){3-5} \cmidrule(lr){6-9}
            & & \multirow{2}{*}{\textit{Yes-or-No}↑} & \multirow{2}{*}{\textit{What}↑} & \multirow{2}{*}{\textit{How}↑} & \multirow{2}{*}{\textit{Distortion}↑} & \multirow{2}{*}{\textit{Other}↑} & \textit{In-context} & \textit{In-context} & \\ 
            & & & & & & & \textit{Distortion}↑ & \textit{Other}↑ & \\
            \midrule
            Random guess & - & 50.00 & 27.86 & 33.31 & 37.89 & 38.48 & 38.28 & 35.82 & 37.80 \\
            \midrule
            GPT-4o \textcolor{gray}{\textit{(2024-08-06)}} & \textcolor{black}{Proprietary Model} & 76.72 & 74.33 & 67.74 & 70.03 & 73.37 & 73.68 & 77.95 & 73.04 \\
            Qwen-VL-Max~\cite{bai2023qwen} & Proprietary Model & 75.60 & 79.43 & 66.09 & 73.39 & 74.08 & 71.00 & 76.92 & 73.63 \\
            Qwen-VL-Plus~\cite{bai2023qwen} & Proprietary Model & 73.77 & 69.47 & 53.88 & 66.21 & 65.72 & 63.81 & 68.75 & 66.04 \\
            Gemini-Pro & Proprietary Model & 68.80 & 73.74 & 62.34 & 66.30 & 71.34 & 63.91 & 73.09 & 68.16 \\
            \midrule
            Emu2-Chat~\cite{sun2024generativemultimodal} & LLaMA-2-33B & 71.81 & 67.25 & 56.18 & 64.78 & 63.19 & 63.48 & 72.24 & 65.28 \\
            mPLUG-Owl2~\cite{ye2023mplugowl2} & LLaMA-2-7B & 72.18 & 57.96 & 56.19 & 56.68 & 69.21 & 53.29 & 72.65 & 61.61 \\  
            LLAMA-Adapter-V2~\cite{gao2023llamaadapterv2} & LLaMA-2-7B & 66.18 & 59.29 & 52.13 & 57.39 & 56.25 & 63.16 & 64.90 & 59.46 \\
            IDEFICS-Instruct~\cite{alayrac2022flamingo} & LLaMA-2-7B & 56.18 & 44.69 & 44.02 & 42.80 & 54.17 & 44.74 & 56.33 & 48.70 \\
            SPHINX~\cite{lin2024sphinx} & LLaMA-2 & 74.18 & 68.81 & 62.07 & 63.62 & 71.76 & 66.12 & 76.33 & 68.56 \\
            InternLM-XComposer-VL~\cite{zhang2023internlm} & InternLM & 69.45 & 65.27 & 60.85 & 61.67 & 70.14 & 56.91 & 75.10 & 65.35 \\
            Qwen-VL~\cite{bai2023qwen} & QwenLM & 63.09 & 58.19 & 56.39 & 50.58 & 62.73 & 57.89 & 73.88 & 59.40 \\
            Otter-v1~\cite{li2023otter} & MPT-7B & 57.09 & 40.71 & 39.55 & 42.22 & 49.31 & 44.08 & 52.65 & 46.35 \\
            LLAVA-v1.5~\cite{liu2023improved} & Vicuna-v1.5-7B & 66.36 & 58.19 & 50.51 & 49.42 & 65.74 & 54.61 & 70.61 & 58.66 \\
            InstructBLIP~\cite{dai2023instructblip} & Vicuna-v1.5-7B & 71.64 & 52.65 & 43.81 & 48.64 & 62.50 & 55.59 & 64.90 & 56.72 \\
            Shikra~\cite{chen2023shikra} & Vicuna-v1.5-7B & 65.64 & 47.35 & 49.09 & 48.83 & 59.49 & 50.00 & 64.08 & 54.65 \\
            MiniGPT-4~\cite{zhu2023minigpt4} & Vicuna-v1.5-13B & 55.82 & 50.22 & 40.37 & 42.02 & 48.38 & 51.97 & 61.22 & 49.03 \\

            \midrule
            AesExpert~\cite{huang2024aesexpert} & Vicuna-v1.5-7B & 73.27 & 64.38 & 53.75 & 70.03 & 73.38 & 73.68 & 77.96 & 64.15 \\

            Q-Instruct~\cite{wu2024qinstruct} & Vicuna-v1.5-7B & 76.18 & 66.37 & 57.61 & 65.18 & 67.59 & 73.06 & 71.53 & 67.09 \\
            
            \rowcolor[gray]{0.90} % Sets a light gray background for the row
            PhotoEye (Ours) & Vicuna-v1.5-7B & \textbf{80.01} & \textbf{76.10} & \textbf{67.02} & \textbf{74.32} & \textbf{74.59} & \textbf{77.30} & \textbf{81.22} & \textbf{74.50} \\
            \bottomrule
        \end{tabular}
    }
    \vspace{-2mm}
    \caption{Model Performance on Q-Bench (LLVisionQA-dev). PhotoEye has a clear advantage. Best open-source model results in \textbf{bold}. \textcolor{black}{Proprietary model results are quoted from Q-Bench~\cite{wu2024qbench}.}}
    \vspace{-2mm}
    \label{tab:tab:qbench}
\end{table*}
PhotoEye clearly outperforms existing open-source models, including both general-purpose models and those specifically fine-tuned on low-level vision and image aesthetics datasets, such as Q-Instruct~\cite{wu2024qinstruct} and AesExpert~\cite{huang2024aesexpert}. Compared to closed-source models, our lightweight approach also demonstrates competitive performance, slightly surpassing GPT-4o.

\begin{table*}[t]
    \centering
    \footnotesize
    \renewcommand{\arraystretch}{0.9}    % Row spacing adjustment
    \setlength{\tabcolsep}{11pt}         % Column spacing adjustment
    \scalebox{0.75}{                     % Scale table to 60% of text width
        \begin{tabular}{llcccccc}    
            \toprule
            \textbf{Model} & \textbf{LLM Backbone} & \textit{Composition}↑ & \textit{Equipments}↑ & \textit{Contrast}↑ & \textit{Techniques}↑ & \textit{Color and Tone}↑ & \textit{Lighting}↑ \\
            \midrule
            GPT-4o \textcolor{gray}{\textit{(2024-08-06)}} & Proprietary Model & 68.11 & 60.24 & 59.09 & 62.75 & 63.97 & 63.74  \\
            \midrule
            LLaVA-v1.6-34B~\cite{liu2023improved} & Hermes-2-Yi-34B & 47.69 & 52.56 & 56.82 & 49.33 & 65.44 & 53.80 \\
            UNIAA~\cite{zhou2024uniaa} & Vicuna-v1.5-7B & 29.89 & 38.61 & 45.45 & 39.56 & 40.19 & 29.24 \\
            AesExpert~\cite{huang2024aesexpert} & Vicuna-v1.5-7B & 51.49 & 60.24 & 56.82 & 53.99 & 61.03 & 64.33  \\
            Q-Instruct~\cite{wu2024qinstruct} & Vicuna-v1.5-7B & 31.02 & 34.77 & 40.91 & 31.37 & 54.38 & 44.44  \\
            \rowcolor[gray]{0.90} % Sets a light gray background for the row
            PhotoEye (Ours) & Vicuna-v1.5-7B & \textbf{68.32} & \textbf{61.39} & \textbf{72.97} & \textbf{65.69} & \textbf{76.00} & \textbf{77.78} \\

            \midrule
            \midrule

            \textbf{Model} & \textbf{LLM Backbone} & \textit{Exposure}↑ & \textit{Post-Processing}↑ & \textit{Aperture and Focus}↑ & \textit{Storytelling}↑ & \textit{Sharpness and Clarity}↑ & \textbf{Overall}↑ \\
            \midrule
            GPT-4o \textcolor{gray}{\textit{(2024-08-06)}} & Proprietary Model & 54.48 & 58.57 & 59.87 & 69.44 & \textbf{61.90} & 64.12 \\
            \midrule
            LLaVA-v1.6-34B~\cite{liu2023improved} & Hermes-2-Yi-34B & 56.55 & 53.33 & 52.87 & 66.67 & 28.57 & 55.68 \\
            UNIAA~\cite{zhou2024uniaa} & Vicuna-v1.5-7B & 25.52 & 38.57 & 40.13 & 30.56 & 23.81 & 36.87 \\
            AesExpert~\cite{huang2024aesexpert} & Vicuna-v1.5-7B & 46.21 & 61.90 & 52.87 & \textbf{72.97} & 33.32 & 60.01 \\
            Q-Instructt~\cite{wu2024qinstruct} & Vicuna-v1.5-7B & 46.21 & 46.19 & 38.85 & 55.56 & 19.05 & 43.86 \\
            \rowcolor[gray]{0.90} % Sets a light gray background for the row
            PhotoEye (Ours) & Vicuna-v1.5-7B & \textbf{69.66} & \textbf{80.95} & \textbf{70.70} & \textbf{75.00} & \textbf{61.90} & \textbf{73.92} \\
            
            \bottomrule
        \end{tabular}
    }
     \vspace{-2mm}
    \caption{Model Performance on PhotoBench Metrics. Best results in \textbf{bold}.}
    \vspace{-4mm}
    \label{tab:photoeye}
\end{table*}

In Tab.~\ref{tab:photoeye}, we compare our model with most competing baselines on PhotoBench, including UNIAA~\cite{zhou2024uniaa}, AesExpert~\cite{huang2024aesexpert} and Q-Instruct~\cite{wu2024qinstruct}, which specially focus on low-level vision and aesthetics.
Note that when reporting results in Tab.~\ref{tab:photoeye}, we merge similar topics into a single category. Categories may overlap - for instance, a question about editing 
 exposure would be counted in both \textit{Post-Processing} and \textit{Exposure}.
PhotoEye notably outperforms baselines. As discussed in Section 5, PhotoBench shows substantially higher expertise and diversity than existing benchmarks, demanding stronger aesthetic understanding capabilities from models. Therefore, PhotoEye, which learns from extensive discussions by photography professionals and enthusiasts, holds a clear advantage.

\vspace{1mm}
\noindent \textbf{Ablations.} We show the effectiveness of our proposed vision fusor and dataset PhotoCritique in Tab.~\ref{tab:ablation_study}. Both contribute to the model. More ablations including qualitative results are available in Appendix.

\begin{table}[t]
   \centering
   \footnotesize
   \renewcommand{\arraystretch}{0.8}   
   \setlength{\tabcolsep}{12pt}         
   \vspace{-0mm}
   \scalebox{0.75}{  
   \begin{tabular}{lcc}
       \toprule
       \textbf{Method} & Q-Bench & PhotoEye \\
       \midrule
       PhotoEye (full method) & \textbf{74.50} & \textbf{73.92} \\
       \hspace{0.1cm} w/o Multi-view Vision Fusor & 70.08 & 68.83 \\
       \hspace{0.1cm} w/o Multi-view Vision Fusor and dataset  & 58.04 & 33.74 \\
       \bottomrule
   \end{tabular}
   }  
   \vspace{-2mm}
   \caption{Ablation Results.}
   \label{tab:ablation_study}
  \vspace{-7mm}
\end{table}

\vspace{1mm}
\noindent \textbf{Qualitative Evaluation.} Providing aesthetic recommendations in conversations is actually the primary need in real-world scenarios. In addition to examples in Fig~\ref{fig:head}, \textcolor{black}{We show examples in Appendix E.} with diverse images and contexts.

\begin{figure}[t]
\vspace{-0mm}
\centerline{\includegraphics[scale=0.25]{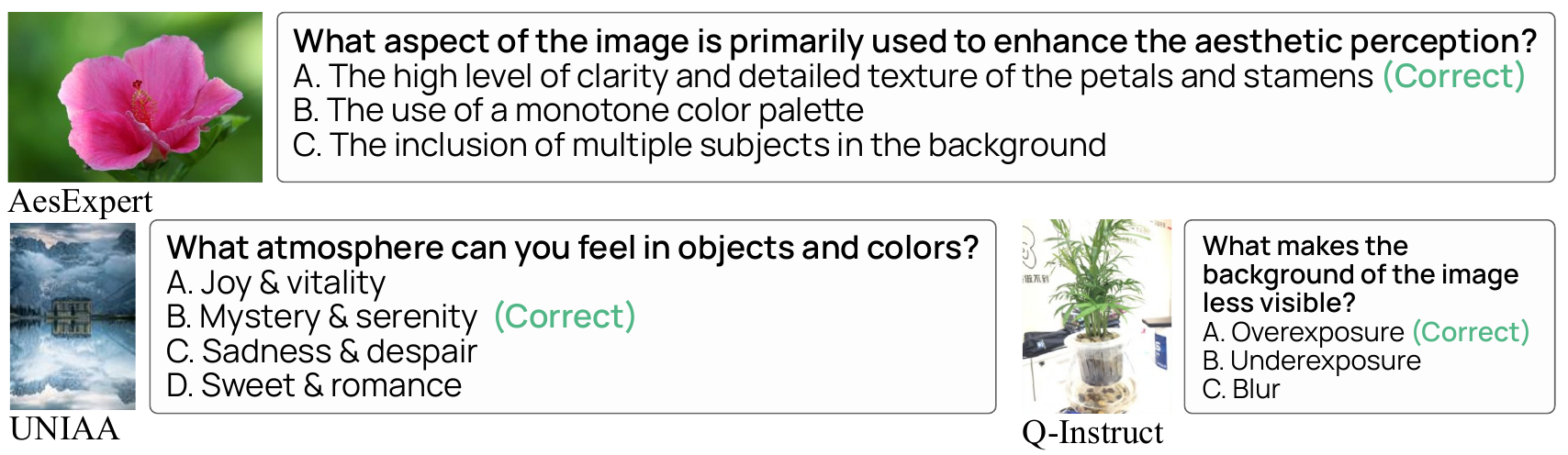}}
\vspace{-2mm}
\caption{Examples in related works from their original papers.}
\label{fig:bench_base}
\vspace{-6mm}
\end{figure}

\subsection{See through PhotoEye}
\begin{figure}[t]
\vspace{-0mm}
\centerline{\includegraphics[scale=0.065]{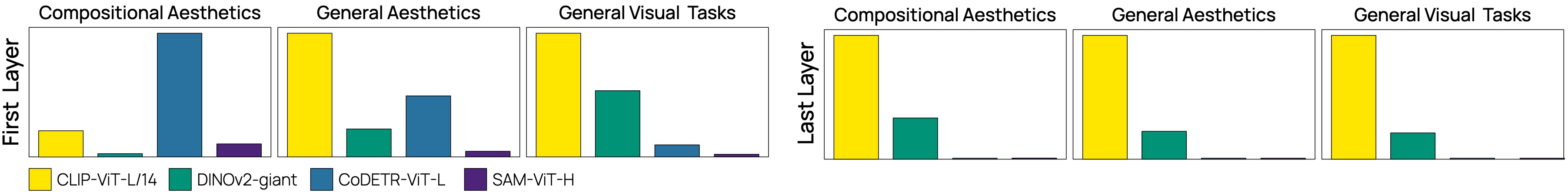}}
\vspace{-1mm}
\caption{Weights of vision encoders across different layers.}
\label{fig:layer}
\vspace{-2mm}
\end{figure}

As multiple visual features from different vision encoders are fused in PhotoEye, we are particularly interested in their roles under different cases and how they contribute to our model. Fig.~\ref{fig:layer} presents the normalized weights of each vision encoder in PhotoEye. We first randomly collect 100 samples from PhotoCritique on compositional aesthetics (e.g., photographic composition, distraction, framing, etc.) and general aesthetics (e.g., storytelling, lighting, etc), respectively, as well as 100 samples for general visual tasks from LLaVA~\cite{liu2023improved} training data. Then we study the averaged weight distribution (normalized) of encoders on different data. 
We have an interesting observation: \textit{(1) aesthetic features are extracted in the first layers, and general features are extracted in the last layers.} Fig.~\ref{fig:layer} shows CLIP and DINO features are dominant regardless of domains in the last layer, which is intuitive as the recognition of objects lays the foundation in both domains. In contrast, aesthetics-related features are primarily extracted from the first layer, especially in compositional problems, where the localization or arrangement of objects are critical. In this case, CoDETR (pre-trained with \textit{detection} objectives) is dominant. Tab.~\ref{tab:com} validated it, where we only activated one encoder during inference to detect their effectiveness in photographic composition.

\vspace{1mm} 
\noindent \textbf{Case Study.} We show how different choices of vision encoders affect model output in real-world conversation scenarios. We provide interesting examples in Appendix F.

\vspace{1mm} 
\noindent \textbf{Learning of Aesthetic Concepts.} Fig.~\ref{fig:eye_intro} demonstrates how PhotoEye effectively learns discriminative aesthetics-related visual representations. In Appendix B, we conduct a deeper investigation into the learning of aesthetic concepts at feature level, which lays the foundation of high-quality aesthetic response from LLM.
\begin{table}[t]
   \centering
   \footnotesize
   \renewcommand{\arraystretch}{0.8}   
   \setlength{\tabcolsep}{5pt}         
   % \vspace{-2mm}
   \scalebox{0.75}{  
   \begin{tabular}{lcccc}
       \toprule
         \textbf{Topic} & CLIP-ViT-L/14 & DINOv2-giant & CoDETR-ViT-L & SAM-ViT-H  \\
       \midrule
       \textit{Composition} & 62.61 & 61.61 & \textbf{64.69} & 60.24  \\ %
       \bottomrule
   \end{tabular}
   }  
   \vspace{-2mm}
   \caption{PhotoEye on \textit{composition} with different encoders.}
   \vspace{-6mm}
   \label{tab:com}
\end{table}

%% file: sec/7_conclusion.tex
\section{Conclusion}
We present PhotoCritique dataset, and PhotoEye, an MLLM with language-guided multi-view vision fusion for aesthetic visual understanding, along with the new benchmark PhotoBench. Extensive experiments show PhotoEye's clear advantage over existing models, advancing aesthetic visual perception in MLLMs.

%% file: main.bbl
\begin{thebibliography}{41}
\providecommand{\natexlab}[1]{#1}
\providecommand{\url}[1]{\texttt{#1}}
\expandafter\ifx\csname urlstyle\endcsname\relax
  \providecommand{\doi}[1]{doi: #1}\else
  \providecommand{\doi}{doi: \begingroup \urlstyle{rm}\Url}\fi

\bibitem[Alayrac et~al.(2022)Alayrac, Donahue, Luc, Miech, Barr, Hasson, Lenc, Mensch, Millican, Reynolds, et~al.]{alayrac2022flamingo}
Jean-Baptiste Alayrac, Jeff Donahue, Pauline Luc, Antoine Miech, Iain Barr, Yana Hasson, Karel Lenc, Arthur Mensch, Katherine Millican, Malcolm Reynolds, et~al.
\newblock Flamingo: a visual language model for few-shot learning.
\newblock \emph{Advances in neural information processing systems}, 2022.

\bibitem[Bai et~al.(2023)Bai, Bai, Yang, Wang, Tan, Wang, Lin, Zhou, and Zhou]{bai2023qwen}
Jinze Bai, Shuai Bai, Shusheng Yang, Shijie Wang, Sinan Tan, Peng Wang, Junyang Lin, Chang Zhou, and Jingren Zhou.
\newblock Qwen-vl: A frontier large vision-language model with versatile abilities.
\newblock \emph{arXiv preprint arXiv:2308.12966}, 2023.

\bibitem[Chen et~al.(2023{\natexlab{a}})Chen, Zhu, Shen, Li, Liu, Zhang, Krishnamoorthi, Chandra, Xiong, and Elhoseiny]{chen2023minigpt}
Jun Chen, Deyao Zhu, Xiaoqian Shen, Xiang Li, Zechun Liu, Pengchuan Zhang, Raghuraman Krishnamoorthi, Vikas Chandra, Yunyang Xiong, and Mohamed Elhoseiny.
\newblock Minigpt-v2: large language model as a unified interface for vision-language multi-task learning.
\newblock \emph{arXiv preprint arXiv:2310.09478}, 2023{\natexlab{a}}.

\bibitem[Chen et~al.(2023{\natexlab{b}})Chen, Zhang, Zeng, Zhang, Zhu, and Zhao]{chen2023shikra}
Keqin Chen, Zhao Zhang, Weili Zeng, Richong Zhang, Feng Zhu, and Rui Zhao.
\newblock Shikra: Unleashing multimodal llm's referential dialogue magic.
\newblock \emph{arXiv preprint arXiv:2306.15195}, 2023{\natexlab{b}}.

\bibitem[Chen et~al.(2023{\natexlab{c}})Chen, Wu, Wang, Su, Chen, Xing, Zhong, Zhang, Zhu, Lu, Li, Luo, Lu, Qiao, and Dai]{chen2023internvl}
Zhe Chen, Jiannan Wu, Wenhai Wang, Weijie Su, Guo Chen, Sen Xing, Muyan Zhong, Qinglong Zhang, Xizhou Zhu, Lewei Lu, Bin Li, Ping Luo, Tong Lu, Yu Qiao, and Jifeng Dai.
\newblock Internvl: Scaling up vision foundation models and aligning for generic visual-linguistic tasks.
\newblock \emph{arXiv preprint arXiv:2312.14238}, 2023{\natexlab{c}}.

\bibitem[Dai et~al.(2023)Dai, Li, Li, Tiong, Zhao, Wang, Li, Fung, and Hoi]{dai2023instructblip}
Wenliang Dai, Junnan Li, Dongxu Li, Anthony Meng~Huat Tiong, Junqi Zhao, Weisheng Wang, Boyang Li, Pascale Fung, and Steven Hoi.
\newblock Instructblip: Towards general-purpose vision-language models with instruction tuning, 2023.

\bibitem[Gao et~al.(2023)Gao, Han, Zhang, Lin, Geng, Zhou, Zhang, Lu, He, Yue, Li, and Qiao]{gao2023llamaadapterv2}
Peng Gao, Jiaming Han, Renrui Zhang, Ziyi Lin, Shijie Geng, Aojun Zhou, Wei Zhang, Pan Lu, Conghui He, Xiangyu Yue, Hongsheng Li, and Yu Qiao.
\newblock Llama-adapter v2: Parameter-efficient visual instruction model.
\newblock \emph{arXiv preprint arXiv:2311.07575}, 2023.

\bibitem[Huang et~al.(2024{\natexlab{a}})Huang, Sheng, Yang, Yuan, Duan, Chen, Li, Lin, and Shi]{huang2024aesexpert}
Yipo Huang, Xiangfei Sheng, Zhichao Yang, Quan Yuan, Zhichao Duan, Pengfei Chen, Leida Li, Weisi Lin, and Guangming Shi.
\newblock {AesExpert}: Towards multi-modality foundation model for image aesthetics perception.
\newblock \emph{arXiv preprint arXiv:2404.09624}, 2024{\natexlab{a}}.

\bibitem[Huang et~al.(2024{\natexlab{b}})Huang, Yuan, Sheng, Yang, Wu, Chen, Yang, Li, and Lin]{huang2024aesbench}
Yipo Huang, Quan Yuan, Xiangfei Sheng, Zhichao Yang, Haoning Wu, Pengfei Chen, Yuzhe Yang, Leida Li, and Weisi Lin.
\newblock {AesBench}: An expert benchmark for multimodal large language models on image aesthetics perception.
\newblock \emph{arXiv preprint arXiv:2401.08276}, 2024{\natexlab{b}}.

\bibitem[Huang et~al.(2024{\natexlab{c}})Huang, Zhang, Lu, Zha, Chen, and Guo]{huang2024visualcritic}
Zhipeng Huang, Zhizheng Zhang, Yiting Lu, Zheng-Jun Zha, Zhibo Chen, and Baining Guo.
\newblock Visualcritic: Making lmms perceive visual quality like humans.
\newblock \emph{arXiv preprint arXiv:2403.12806}, 2024{\natexlab{c}}.

\bibitem[Kirillov et~al.(2023)Kirillov, Mintun, Ravi, Mao, Rolland, Gustafson, Xiao, Whitehead, Berg, Lo, et~al.]{kirillov2023segment}
Alexander Kirillov, Eric Mintun, Nikhila Ravi, Hanzi Mao, Chloe Rolland, Laura Gustafson, Tete Xiao, Spencer Whitehead, Alexander~C Berg, Wan-Yen Lo, et~al.
\newblock Segment anything.
\newblock \emph{arXiv preprint arXiv:2304.02643}, 2023.

\bibitem[Li et~al.(2023)Li, Zhang, Chen, Wang, Yang, and Liu]{li2023otter}
Bo Li, Yuanhan Zhang, Liangyu Chen, Jinghao Wang, Jingkang Yang, and Ziwei Liu.
\newblock Otter: A multi-modal model with in-context instruction tuning.
\newblock \emph{arXiv preprint arXiv:2305.03726}, 2023.

\bibitem[Lin et~al.(2023)Lin, Liu, Zhang, Gao, Qiu, Xiao, Qiu, Lin, Shao, Chen, Han, Huang, Zhang, He, Li, and Qiao]{Lin2023}
Ziyi Lin, Chris Liu, Renrui Zhang, Peng Gao, Longtian Qiu, Han Xiao, Han Qiu, Chen Lin, Wenqi Shao, Keqin Chen, Jiaming Han, Siyuan Huang, Yichi Zhang, Xuming He, Hongsheng Li, and Yu Qiao.
\newblock {SPHINX: The Joint Mixing of Weights, Tasks, and Visual Embeddings for Multi-modal Large Language Models}.
\newblock \emph{arXiv preprint arXiv:2311.07575}, 2023.

\bibitem[Lin et~al.(2024)Lin, Liu, Zhang, Gao, Qiu, Xiao, Qiu, Lin, Shao, Chen, Han, Huang, Zhang, He, Li, and Qiao]{lin2024sphinx}
Ziyi Lin, Chris Liu, Renrui Zhang, Peng Gao, Longtian Qiu, Han Xiao, Han Qiu, Chen Lin, Wenqi Shao, Keqin Chen, Jiaming Han, Siyuan Huang, Yichi Zhang, Xuming He, Hongsheng Li, and Yu Qiao.
\newblock Sphinx: The joint mixing of weights, tasks, and visual embeddings for multi-modal large language models.
\newblock \emph{arXiv preprint arXiv:2311.07575}, 2024.

\bibitem[Liu et~al.(2023)Liu, Li, Li, and Lee]{liu2023improved}
Haotian Liu, Chunyuan Li, Yuheng Li, and Yong~Jae Lee.
\newblock Improved baselines with visual instruction tuning.
\newblock \emph{arXiv preprint arXiv:2310.03744}, 2023.

\bibitem[Liu et~al.(2024)Liu, Li, Wu, and Lee]{liu2024visual}
Haotian Liu, Chunyuan Li, Qingyang Wu, and Yong~Jae Lee.
\newblock Visual instruction tuning.
\newblock \emph{Advances in neural information processing systems}, 36, 2024.

\bibitem[Murray et~al.(2012)Murray, Marchesotti, and Perronnin]{murray2012ava}
Naila Murray, Luca Marchesotti, and Florent Perronnin.
\newblock Ava: A large-scale database for aesthetic visual analysis.
\newblock In \emph{2012 IEEE Conference on Computer Vision and Pattern Recognition}, pages 2408--2415. IEEE, 2012.

\bibitem[Nieto et~al.(2022)Nieto, Celona, and Fernandez-Labrador]{nieto2022understanding}
Daniel~Vera Nieto, Luigi Celona, and Clara Fernandez-Labrador.
\newblock Understanding aesthetics with language: A photo critique dataset for aesthetic assessment.
\newblock In \emph{NeurIPS Track on Datasets and Benchmarks}, 2022.

\bibitem[OpenAI(2023)]{OpenAI2023GPT4}
OpenAI.
\newblock Gpt-4 technical report.
\newblock 2023.

\bibitem[Oquab et~al.(2023)Oquab, Darcet, Moutakanni, Vo, Szafraniec, Khalidov, Fernandez, Haziza, Massa, El-Nouby, et~al.]{oquab2023dinov2}
Maxime Oquab, Timothée Darcet, Théo Moutakanni, Huy Vo, Marc Szafraniec, Vasil Khalidov, Pierre Fernandez, Daniel Haziza, Francisco Massa, Alaaeldin El-Nouby, et~al.
\newblock Dinov2: Learning robust visual features without supervision.
\newblock \emph{arXiv preprint arXiv:2304.07193}, 2023.

\bibitem[Peng et~al.(2023)Peng, Wang, Dong, Hao, Huang, Ma, and Wei]{peng2023kosmos}
Zhiliang Peng, Wenhui Wang, Li Dong, Yaru Hao, Shaohan Huang, Shuming Ma, and Furu Wei.
\newblock Kosmos-2: Grounding multimodal large language models to the world.
\newblock \emph{arXiv preprint arXiv:2306.14824}, 2023.

\bibitem[Qi et~al.(2024{\natexlab{a}})Qi, Zhao, and Li]{qi2024easy}
Daiqing Qi, Handong Zhao, and Sheng Li.
\newblock Easy regional contrastive learning of expressive fashion representations.
\newblock \emph{Advances in Neural Information Processing Systems}, 37:\penalty0 20480--20509, 2024{\natexlab{a}}.

\bibitem[Qi et~al.(2024{\natexlab{b}})Qi, Zhao, Wei, and Li]{qi2024tag}
Daiqing Qi, Handong Zhao, Zijun Wei, and Sheng Li.
\newblock Tag-grounded visual instruction tuning with retrieval augmentation.
\newblock In \emph{Proceedings of the 2024 Conference on Empirical Methods in Natural Language Processing}, pages 2008--2026, 2024{\natexlab{b}}.

\bibitem[Qi et~al.(2024{\natexlab{c}})Qi, Zhao, Zhang, and Li]{qi2024generalizing}
Daiqing Qi, Handong Zhao, Aidong Zhang, and Sheng Li.
\newblock Generalizing to unseen domains via text-guided augmentation: A training-free approach.
\newblock In \emph{European Conference on Computer Vision}, pages 285--300. Springer, 2024{\natexlab{c}}.

\bibitem[Radford et~al.(2021)Radford, Kim, Hallacy, Ramesh, Goh, Agarwal, Sastry, Askell, Mishkin, Clark, et~al.]{radford2021learning}
Alec Radford, Jong~Wook Kim, Chris Hallacy, Aditya Ramesh, Gabriel Goh, Sandhini Agarwal, Girish Sastry, Amanda Askell, Pamela Mishkin, Jack Clark, et~al.
\newblock Learning transferable visual models from natural language supervision.
\newblock In \emph{International conference on machine learning}, pages 8748--8763. PMLR, 2021.

\bibitem[Ren et~al.(2017)Ren, Shen, Lin, Mech, and Foran]{ren2017personalized}
Jian Ren, Xiaohui Shen, Zhe Lin, Radomir Mech, and David~J. Foran.
\newblock Personalized image aesthetics.
\newblock In \emph{The IEEE International Conference on Computer Vision (ICCV)}, 2017.

\bibitem[Sun et~al.(2024)Sun, Cui, Zhang, Zhang, Yu, Luo, Wang, Rao, Liu, Huang, and Wang]{sun2024generativemultimodal}
Quan Sun, Yufeng Cui, Xiaosong Zhang, Fan Zhang, Qiying Yu, Zhengxiong Luo, Yueze Wang, Yongming Rao, Jingjing Liu, Tiejun Huang, and Xinlong Wang.
\newblock Generative multimodal models are in-context learners.
\newblock \emph{arXiv preprint arXiv:2312.13286}, 2024.

\bibitem[Tong et~al.(2024)Tong, Brown, Wu, Woo, Middepogu, Akula, Yang, Yang, Iyer, Pan, Wang, Fergus, LeCun, and Xie]{tong2024cambrian1fullyopenvisioncentric}
Shengbang Tong, Ellis Brown, Penghao Wu, Sanghyun Woo, Manoj Middepogu, Sai~Charitha Akula, Jihan Yang, Shusheng Yang, Adithya Iyer, Xichen Pan, Austin Wang, Rob Fergus, Yann LeCun, and Saining Xie.
\newblock Cambrian-1: A fully open, vision-centric exploration of multimodal llms, 2024.

\bibitem[Wu et~al.(2024{\natexlab{a}})Wu, Zhang, Zhang, Chen, Liao, Wang, Li, Sun, Yan, Zhai, and Lin]{wu2024qbench}
Haoning Wu, Zicheng Zhang, Erli Zhang, Chaofeng Chen, Liang Liao, Annan Wang, Chunyi Li, Wenxiu Sun, Qiong Yan, Guangtao Zhai, and Weisi Lin.
\newblock {Q-BENCH}: A benchmark for general-purpose foundation models on low-level vision.
\newblock In \emph{Proceedings of the International Conference on Learning Representations (ICLR)}, 2024{\natexlab{a}}.

\bibitem[Wu et~al.(2024{\natexlab{b}})Wu, Zhang, Zhang, Chen, Liao, Wang, Xu, Li, Hou, Zhai, Xue, Sun, Yan, and Lin]{wu2024qinstruct}
Haoning Wu, Zicheng Zhang, Erli Zhang, Chaofeng Chen, Liang Liao, Annan Wang, Kaixin Xu, Chunyi Li, Jingwen Hou, Guangtao Zhai, Geng Xue, Wenxiu Sun, Qiong Yan, and Weisi Lin.
\newblock {Q-Instruct}: Improving low-level visual abilities for multi-modality foundation models.
\newblock In \emph{Proceedings of the IEEE/CVF Conference on Computer Vision and Pattern Recognition (CVPR)}, 2024{\natexlab{b}}.

\bibitem[Wu et~al.(2024{\natexlab{c}})Wu, Zhu, Zhang, Zhang, Chen, Liao, Li, Wang, Sun, Yan, Liu, Zhai, Wang, and Lin]{wu2024openended}
Haoning Wu, Hanwei Zhu, Zicheng Zhang, Erli Zhang, Chaofeng Chen, Liang Liao, Chunyi Li, Annan Wang, Wenxiu Sun, Qiong Yan, Xiaohong Liu, Guangtao Zhai, Shiqi Wang, and Weisi Lin.
\newblock Towards open-ended visual quality comparison.
\newblock In \emph{Proceedings of the European Conference on Computer Vision (ECCV)}, 2024{\natexlab{c}}.

\bibitem[Yang et~al.(2022)Yang, Xu, Li, Qie, Li, Zhang, and Guo]{yang2022personalized}
Yuzhe Yang, Liwu Xu, Leida Li, Nan Qie, Yaqian Li, Peng Zhang, and Yandong Guo.
\newblock Personalized image aesthetics assessment with rich attributes.
\newblock In \emph{Proceedings of the IEEE/CVF Conference on Computer Vision and Pattern Recognition (CVPR)}, pages 19861--19869, 2022.

\bibitem[Ye et~al.(2023{\natexlab{a}})Ye, Xu, Xu, Ye, Yan, Zhou, Wang, Hu, Shi, Shi, et~al.]{ye2023mplug}
Qinghao Ye, Haiyang Xu, Guohai Xu, Jiabo Ye, Ming Yan, Yiyang Zhou, Junyang Wang, Anwen Hu, Pengcheng Shi, Yaya Shi, et~al.
\newblock mplug-owl: Modularization empowers large language models with multimodality.
\newblock \emph{arXiv preprint arXiv:2304.14178}, 2023{\natexlab{a}}.

\bibitem[Ye et~al.(2023{\natexlab{b}})Ye, Xu, Ye, Yan, Hu, Liu, Qian, Zhang, Huang, and Zhou]{ye2023mplugowl2}
Qinghao Ye, Haiyang Xu, Jiabo Ye, Ming Yan, Anwen Hu, Haowei Liu, Qi Qian, Ji Zhang, Fei Huang, and Jingren Zhou.
\newblock mplug-owl2: Revolutionizing multi-modal large language model with modality collaboration.
\newblock \emph{arXiv preprint arXiv:2311.04257}, 2023{\natexlab{b}}.

\bibitem[You et~al.(2023)You, Zhang, Gan, Du, Zhang, Wang, Cao, Chang, and Yang]{you2023ferret}
Haoxuan You, Haotian Zhang, Zhe Gan, Xianzhi Du, Bowen Zhang, Zirui Wang, Liangliang Cao, Shih-Fu Chang, and Yinfei Yang.
\newblock Ferret: Refer and ground anything anywhere at any granularity.
\newblock \emph{arXiv preprint arXiv:2310.07704}, 2023.

\bibitem[Zhang et~al.(2023)Zhang, Wang, Cao, Xu, Ouyang, Zhao, Ding, Zhang, Duan, Yan, et~al.]{zhang2023internlm}
Pan Zhang, Xiaoyi Dong~Bin Wang, Yuhang Cao, Chao Xu, Linke Ouyang, Zhiyuan Zhao, Shuangrui Ding, Songyang Zhang, Haodong Duan, Hang Yan, et~al.
\newblock Internlm-xcomposer: A vision-language large model for advanced text-image comprehension and composition.
\newblock \emph{arXiv preprint arXiv:2309.15112}, 2023.

\bibitem[Zhong et~al.(2023)Zhong, Zhou, and Qiu]{zhong2023aesthetically}
Zhipeng Zhong, Fei Zhou, and Guoping Qiu.
\newblock Aesthetically relevant image captioning.
\newblock In \emph{Proceedings of the AAAI Conference on Artificial Intelligence (AAAI)}, 2023.

\bibitem[Zhou et~al.(2024)Zhou, Su, Zheng, Wang, Chen, Yuan, and Zhang]{zhou2024uniaa}
Zhaokun Zhou, Yiwei Su, Amin Zheng, Qiulin Wang, Rui Chen, Li Yuan, and Di Zhang.
\newblock {UNIAA}: A unified multi-modal image aesthetic assessment baseline and benchmark.
\newblock \emph{arXiv preprint arXiv:2404.09619}, 2024.

\bibitem[Zhu et~al.(2023)Zhu, Chen, Shen, Li, and Elhoseiny]{zhu2023minigpt4}
Deyao Zhu, Jun Chen, Xiaoqian Shen, Xiang Li, and Mohamed Elhoseiny.
\newblock Minigpt-4: Enhancing vision-language understanding with advanced large language models.
\newblock \emph{arXiv preprint arXiv:2304.10592}, 2023.

\bibitem[Zong et~al.(2023)Zong, Song, and Liu]{zong2023detrs}
Zhuofan Zong, Guanglu Song, and Yu Liu.
\newblock Detrs with collaborative hybrid assignments training.
\newblock In \emph{Proceedings of the IEEE/CVF International Conference on Computer Vision}, pages 6748--6758, 2023.

\bibitem[Zong et~al.(2024)Zong, Ma, Shen, Song, Shao, Jiang, Li, and Liu]{Zong2024}
Zhuofan Zong, Bingqi Ma, Dazhong Shen, Guanglu Song, Hao Shao, Dongzhi Jiang, Hongsheng Li, and Yu Liu.
\newblock Mova: Adapting mixture of vision experts to multimodal context.
\newblock In \emph{Proceedings of the 38th Conference on Neural Information Processing Systems (NeurIPS)}, 2024.

\end{thebibliography}
